%% file: main.tex
\documentclass[conference]{IEEEtran}

\usepackage[backend=biber]{biblatex}
\AtEveryBibitem{%
  \clearfield{urlyear}
  \clearfield{doi}
  \clearfield{url}
  \clearfield{note}
  \clearlist{location}
  \clearname{editor}
  \clearfield{pages}
  \clearfield{issn}
}

\usepackage{amsmath}
\usepackage{graphicx}
\usepackage{textcomp}
\usepackage{svg}
\usepackage{xcolor}
\usepackage{pifont}
\definecolor{others}{RGB}{0, 0, 0}
\definecolor{barrier}{RGB}{144, 128, 112}
\definecolor{bicycle}{RGB}{100, 230, 245}
\definecolor{bus}{RGB}{100, 80, 250}
\definecolor{car}{RGB}{100, 150, 245}
\definecolor{construction_vehicle}{RGB}{0, 0, 255}
\definecolor{motorcycle}{RGB}{30, 60, 150}
\definecolor{pedestrian}{RGB}{250, 30, 30}
\definecolor{traffic_cone}{RGB}{255, 0, 0}
\definecolor{trailer}{RGB}{0, 140, 255}
\definecolor{truck}{RGB}{80, 30, 180}
\definecolor{driveable_surface}{RGB}{255, 0, 255}
\definecolor{other_flat}{RGB}{75, 0, 175}
\definecolor{sidewalk}{RGB}{75, 0, 75}
\definecolor{terrain}{RGB}{60, 180, 112}
\definecolor{manmade}{RGB}{255, 150, 0}
\definecolor{vegetation}{RGB}{0, 175, 0}

\begin{document}

\title{4DRC-OCC: Robust Semantic Occupancy Prediction Through Fusion of 4D Radar and Camera}

\author{\IEEEauthorblockN{David Ninfa}
\IEEEauthorblockA{Cognitive Robotics\\Delft University of Technology}
\and
\IEEEauthorblockN{Dr. Andras Palffy}
\IEEEauthorblockA{Perciv AI}
\and
\IEEEauthorblockN{Dr. Holger Caesar}
\IEEEauthorblockA{Cognitive Robotics\\Delft University of Technology}
}

\maketitle
\thispagestyle{plain}
\pagestyle{plain}

\input{abstract}

\IEEEpeerreviewmaketitle

\input{macros}

\input{introduction}
\input{related_work}
\input{method}
\input{experiments}
\input{conclusion}
\input{acknowledgement}

\printbibliography[title=References]

\end{document}

%% file: abstract.tex
\begin{abstract}
Autonomous driving systems require robust and reliable perception across diverse environmental conditions, yet current approaches to 3D semantic occupancy prediction face challenges in adverse weather and lighting. This paper presents the first study on fusing 4D radar and cameras for 3D semantic occupancy prediction. This fusion offers significant advantages for robust and accurate perception as 4D radar provides reliable range, velocity, and angle information, even in challenging conditions, complementing rich camera semantic and texture details.
Additionally, we demonstrate that incorporating depth cues with camera image pixels supports lifting 2D images to 3D, enhancing the accuracy of scene reconstruction. Secondly, we introduce a fully automatically labeled dataset specifically designed for training semantic occupancy models, demonstrating its ability to reduce the need for costly manual annotation substantially. Our results highlight the robustness of 4D radar in a wide range of challenging scenarios, showcasing its potential to advance perception for autonomous vehicles.\\

\end{abstract}

%% file: macros.tex
\newcommand{\scrbaseline}{Baseline}
\newcommand{\ftbaseline}{Baseline-ft}
\newcommand{\scrRO}{Radar Only}
\newcommand{\scrAnoelev}{Version AN}
\newcommand{\scrA}{Version A}
\newcommand{\ftAnoelev}{Version AN-ft}
\newcommand{\ftA}{Version A-ft}
\newcommand{\ftB}{Version B-ft}
\newcommand{\ftC}{Version C-ft}
\newcommand{\scrB}{Version B}
\newcommand{\scrC}{Version C}

%% file: introduction.tex
\section{Introduction}

Accurate 3D scene understanding is essential for autonomous vehicles' safe and reliable operation. Recent advancements in camera-based semantic occupancy prediction have enhanced our ability to generate a detailed 3D representation of a scene from 2D images \cite{xu_3d_2023}. Although these monocular methods can offer rich semantic insights, they face notable challenges, such as handling adverse weather, managing occlusions, and accurately estimating depth when lifting from 2D to 3D. These limitations present significant obstacles in real-world settings, where reliable perception is crucial for effective and safe operation.

\begin{figure}[ht]
    \centering
    \includegraphics[width=\columnwidth]{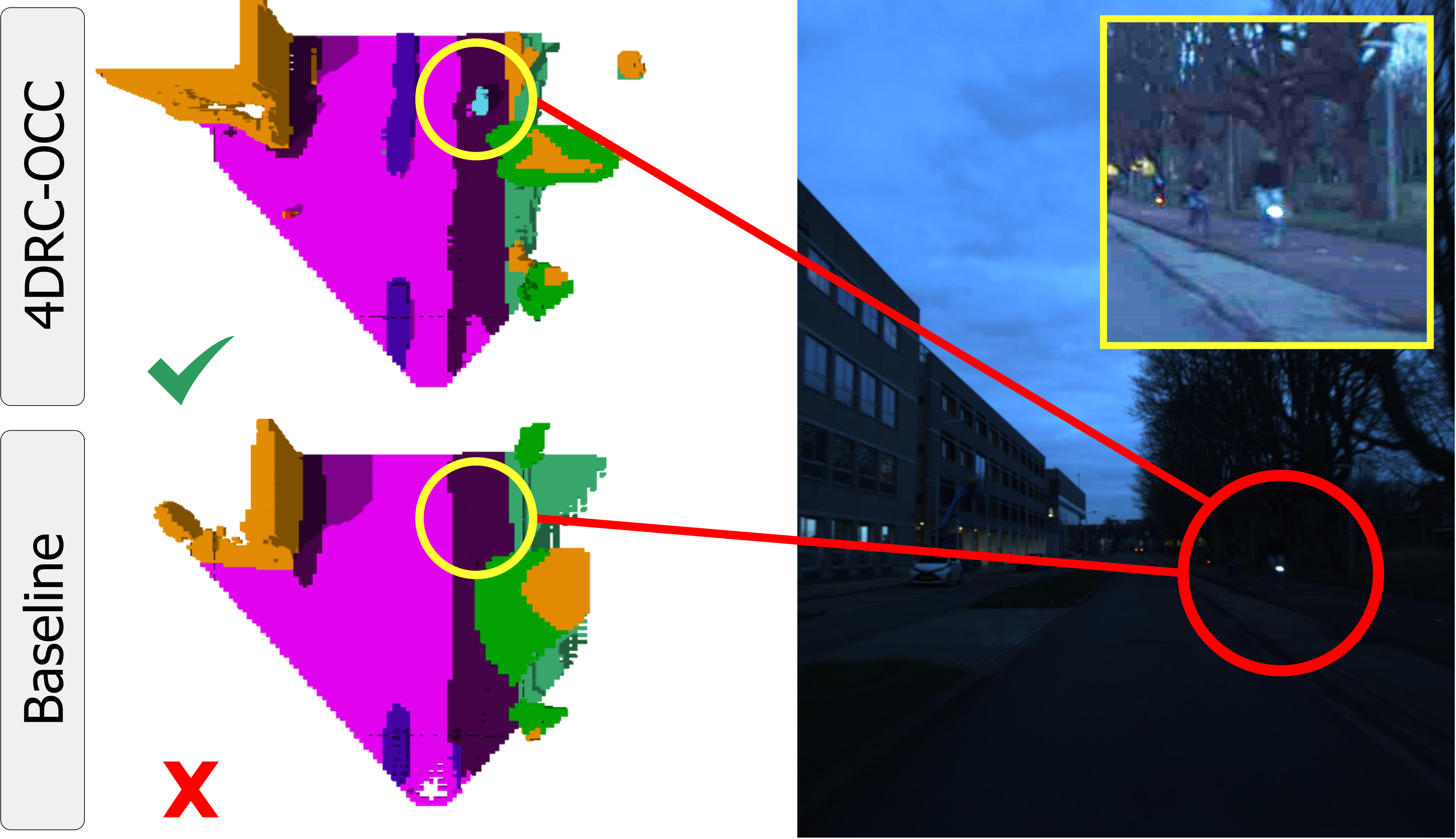}
    \caption{Adverse lighting conditions significantly impact the camera-only baseline; however, our network still detects the cyclist through radar-camera fusion.}
    \label{fig:example0}
\end{figure}

In this work, we introduce 4DRC-OCC, a novel framework for 3D semantic occupancy prediction that leverages 4D radar as a complementary sensor to cameras, marking the first study on the fusion of 4D radar and camera for this purpose. Radar, as a low-frequency active sensor, offers robust depth and velocity information while remaining resilient to adverse weather and lighting conditions, key advantages that address the limitations of camera-only systems \cite{palffy_cnn_2020}. 4D radar surpasses traditional radar by capturing range, velocity, and angular information in azimuth and elevation, providing a detailed 3D representation of the environment. This enhances object localization, discrimination, and overall spatial understanding. Its robustness in adverse weather and lighting conditions, combined with a broader field of view and extended range, makes it a valuable complement to camera data. Additionally, 4D radar generates denser and more informative point clouds, facilitating more effective fusion with visual data for applications like 3D semantic occupancy prediction \cite{fan_4d_2024}. However, integrating radar and camera data presents challenges due to the sparse and noisy nature of radars and limited semantic information. Despite the potential of 4D radar, existing research has primarily focused on camera and LiDAR fusion methods, leaving these unique strengths underutilized in achieving robust scene understanding.

Camera-based methods face inherent limitations in real-world scenarios. They are sensitive to environmental conditions like weather and lighting, often struggling in low-visibility settings such as nighttime, rain, or fog \cite{tan_3-d_2023}, which can compromise system reliability and safety in autonomous applications. Furthermore, vision-based techniques, such as bird’s-eye view (BEV) projections, collapse the vertical dimension, leading to a loss of critical geometric information and making it difficult to accurately capture the 3D structure of objects in complex scenes \cite{li_fb-occ_2023}. These monocular approaches also encounter challenges with depth estimation, limiting their ability to project 2D images into 3D space accurately. This often results in unstable depth predictions and localization errors, particularly with occluded or distant objects \cite{xiong_lxl_2023}. 
Due to their inherent range limitations, the distance at which objects can be reliably detected and recognized is restricted \cite{cai_low-cost_2022}, in contrast to radar, which can detect objects at distances of up to 200 meters \cite{cai_low-cost_2022}. Our method addresses many of these limitations by fusing 4D radar with camera data, enabling a more robust occupancy prediction. However, leveraging radar data entirely remains a significant research challenge due to its data sparsity and noise.

Our 4DRC-OCC approach employs a radar-aided image-lifting mechanism to combine radar and camera data across multiple scales effectively. Unlike conventional sampling methods focusing on discrete depth points, this method distributes image features to neighboring 3D points at various scales. This approach mitigates noise, fills gaps in radar data, and ensures spatial consistency. Camera features are mapped into a voxel-based 3D representation and fused with 4D radar voxel features across different scales, enabling the model to capture fine-grained details and broader spatial contexts. This multi-scale fusion allows the model to capture global context and fine local details, enhancing its overall spatial understanding and feature representation. A key challenge in this domain is the labor-intensive process of creating multi-modal datasets with accurate ground truth labels, which slows progress in radar-based system development. To tackle this issue, we introduce an automatically labeled dataset featuring point-wise annotations, removing the need for manual labeling. This dataset effectively supports model training, demonstrating the practicality of automated labeling and its potential to drive advancements in radar-camera fusion research. In summary, the key contributions of this paper are:

\begin{itemize} 
    \item \textbf{4DRC-OCC}: We propose an architecture for semantic occupancy prediction that fuses 4D radar data with camera features effectively. Specifically, we first fuse radar and lifted camera features to create a comprehensive 3D representation. In addition, we enhance the camera lifting mechanism by incorporating depth cues from radar, addressing the ill-posed problem of depth estimation in monocular occupancy prediction. This is achieved by either adding depth as an additional channel to the image input or generating pseudo-depth images from radar point clouds, which are then concatenated with the camera image features to improve depth information during the feature lifting process.\\
    
    \item \textbf{Auto-labeled Ground Truth}: To address the labor-intensive task of manually annotating large datasets, we construct an automatically labeled dataset with ground truth annotations generated entirely without human intervention. A key contribution of our work is demonstrating that this dataset supports effective training, making it possible to accelerate research in radar-camera fusion for semantic occupancy prediction.\\
\end{itemize}

%% file: related_work.tex
\section{Related Work}

\definecolor{red_}{HTML}{CC0000}
\definecolor{blue_}{HTML}{0066CC}
\begin{figure*}[htb!]
    \centering
    \includegraphics[width=\textwidth]{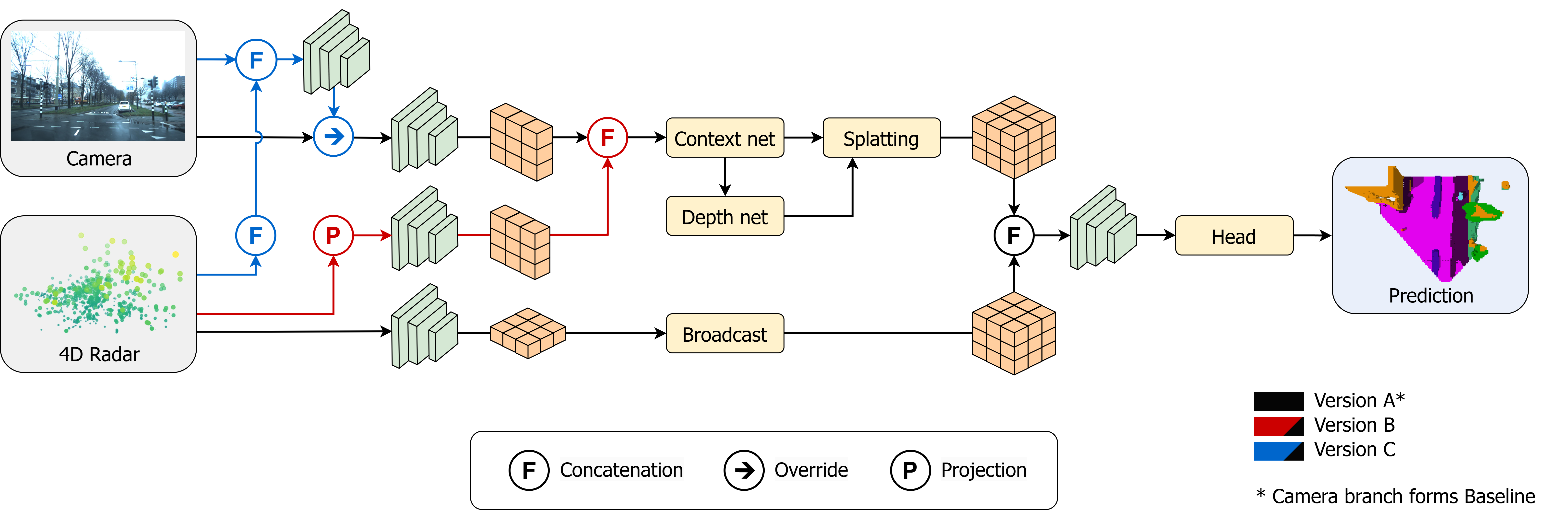}
    \caption{The architecture of 4DRC-OCC: The main network, Version \textbf{A}, processes radar and camera separately in a multi-scale manner, reaching a higher level of abstraction before merging them in voxel space to produce the final prediction. Versions \textcolor{red_}{\textbf{B}} and Version \textcolor{blue_}{\textbf{C}} build upon version \textbf{A} by incorporating additional radar depth cues at different stages, enhancing the camera image and further assisting the lifting mechanism. This allows for a more refined fusion of radar and camera information, improving performance.}
    \label{fig:architecture}
\end{figure*}

\subsection{Monocular Methods}
Networks that only use camera images face challenges due to the absence of direct depth information, which is critical for accurate spatial understanding. To overcome this, feature lifting mechanisms are employed that lift 2D image data into 3D space, where two main approaches are often used to perform this lifting: splatting and sampling.

Splatting-based methods utilize a projection technique where 2D image features are lifted to a bird’s eye view (BEV) or voxel grid through a dense, ray-based projection that assigns depth distributions across all possible 3D locations within the projection range. This dense mapping approach allows the entire 3D scene to be filled with features simultaneously but risks introducing false features due to its less selective nature. A well-known splatting-based network is MonoScene \cite{cao_monoscene_2022}, which uses the Lift-Splat-Shoot (LSS) \cite{vedaldi_lift_2020} paradigm to project 2D image features into a BEV map. In LSS, the extrinsic and intrinsic matrices of the camera are used to compute depth distributions along each pixel projection ray, lifting features onto a BEV plane through depth-aware splatting. BEVDepth \cite{li_bevdepth_2023} enhanced the splatting approach by incorporating lidar-based supervision to improve depth accuracy, showing the impact of precise depth estimation on BEV performance. To optimize feature aggregation, BEVStereo \cite{li_bevstereo_2023} introduced a temporal stereo approach, refining the depth map further and achieving more consistent feature accumulation across time steps. This splatting methodology has evolved, but the reliance on accurate depth estimates and splatting operations introduces training complexities, particularly as depth accuracy significantly impacts overall BEV performance. More recent approaches \cite{li_fb-occ_2023, ming_occfusion_2024, huang_tri-perspective_2023, ma_cotr_2023} have aimed at creating a complete 3D feature volume by voxelizing the pseudo 3D space instead of relying solely on BEV projections, allowing for denser feature representation and increased spatial detail. OccFormer \cite{zhang_occformer_2023} adapts the LSS technique but applies a transformer framework to process long-range dependencies within the scene and combines the BEV splatted features with local-global attention, which helps manage the dense spatial features from splatting while retaining fine-grained details within each local area. FB-OCC \cite{li_fb-occ_2023} proposes a backward view-transformation module to bolster semantic richness in the occupancy map. Multi-Scale Occ \cite{ding_multi-scale_2023} combines multi-scale fusion to implement global and local features, aiming for a robust 3D feature volume. Despite these enhancements, depth ground truth dependency remains a limitation, necessitating additional supervision to refine depth predictions.

Sampling-based methods lift features from a 2D image plane to predefined 3D voxel or grid coordinates by allocating specific sampling points within a continuous 3D space. This approach is beneficial for approximating depth with limited data, as it lifts features along projection rays from known image coordinates to achieve more accurate 3D occupancy representation, with the added advantage of selectively avoiding free space. BEVFormer \cite{li_bevformer_2022} introduced the concept of a BEV query plane, where BEV queries are projected onto the multi-view images using camera intrinsics and extrinsics. This enables feature aggregation without depth lifting, using temporal self-attention to incorporate historical BEV states, thereby enhancing consistency over time. Similarly, DETR3D \cite{wang_detr3d_2021} defines a set of 3D query points that are projected onto images, aggregating features without needing explicit depth estimation. By utilizing transformer mechanisms to gather spatial features, these sampling-based methods reduce the computational demands of dense splatting while still capturing key 3D spatial information. Simple-BEV introduces a bilinear sampling method. Simple-BEV predefines 3D voxel coordinates, and samples 2D image features at each coordinate point. This approach provides a more uniform sampling across distances, preserving feature continuity for near and far objects. The uniformity of sampling in Simple-BEV \cite{harley_simple-bev_2023} is particularly effective in urban scenes where objects can vary widely in distance from the camera. TPVFormer \cite{huang_tri-perspective_2023} adapts the BEV query plane to include three orthogonal planes, capturing the 3D space from multiple perspectives, which SurroundOcc \cite{wei_surroundocc_2023} further enhances by using a 3D query volume for dense sampling across the scene. PanoOcc \cite{wang_panoocc_2023} implements a sparse representation for dense 3D features, optimizing memory and computational load. Although sampling-based approaches are efficient, reliance on transformer layers for multi-view querying can result in high memory consumption and slower training times.

\subsection{Multi-Modal Methods}
To overcome the limitations of monocular methods, multi-modal semantic occupancy networks employ additional sensory inputs that offer complementary spatial data for robust 3D scene reconstruction. HyDRa \cite{wolters_unleashing_2024} combines data from image and radar sensors to improve 3D detection and semantic occupancy. It uses a Height Association Transformer (HAT) to perform cross-attention between radar and image features, producing dense depth information despite the radar’s lack of elevation angle measurement. BEVFusion \cite{liu_bevfusion_2023} merges camera and radar BEV representations, further refined by radar-guided backward projection and task-specific heads, and then processes the fused features. HyDRa’s radar-weighted depth consistency addresses spatial misalignment by enforcing consistency between radar and image depth cues, compensating for the absence of elevation data. OccFusion \cite{ming_occfusion_2024} processes surround-view images, lidar, and radar data, with each modality contributing distinct features. A 2D backbone transforms image features into BEV representations, while a 3D backbone handles lidar and radar point clouds. Dynamic fusion modules merge BEV and 3D feature volumes across scales, followed by global-local attention fusion for refined depth estimation. OccFusion’s radar point cloud employs VoxelNet \cite{zhou_voxelnet_2018} to extract voxel features, and the architecture relies on SENet blocks to emphasize relevant channels, compensating for missing elevation information of conventional radars. These methods are limited by the traditional inability to provide elevation angle measurements. Combining radar with camera-based systems significantly improves performance, especially in long-range scenarios where cameras struggle with depth estimation \cite{wolters_unleashing_2024}.

\subsection{Radar-Camera Fusion}
In radar-camera fusion networks, the fusion mechanism significantly impacts object detection and semantic occupancy prediction efficiency and accuracy \cite{yao_radar-camera_2023}. These networks employ radar and camera data at different stages. Early fusion methods like RRPN \cite{nabati_rrpn_2019} use radar to generate object anchor proposals within the camera’s coordinate system, ensuring accurate spatial calibration but often requiring alignment adjustments due to radar’s limited resolution. CRAFT \cite{kim_craft_2023} leverages radar point clouds to enhance camera-based object proposals using cross-attention layers to capture spatial and contextual information. Feature-level fusion networks, which operate at a middle level of abstraction, are more common in advanced radar-camera fusion. CenterFusion \cite{nabati_centerfusion_2021} exemplifies projection-based feature fusion by associating radar detections with object center points in camera images. It uses a frustum-based approach to match radar points with image objects and augments image features with radar-derived depth and velocity. CRF-Net \cite{nobis_deep_2019} follows a similar method, where radar data is projected onto the image plane and then fused with camera features at multiple layers, emphasizing radar contributions differently depending on each layer’s requirements. BEV-based feature fusion, which uses BEV representations for spatially coherent scene understanding, is prominent in recent networks. For instance, RCBEV \cite{lin_rcbevdet_2024} implements a dual-module fusion strategy: a point-fusion module for voxelized radar features and camera frustums, followed by ROI-fusion to produce BEV heatmaps for object positioning and confidence estimation. SAF-FCOS \cite{chang_spatial_2020} further incorporates a Spatial Attention Fusion (SAF) module to weigh radar data relevance spatially, emphasizing regions where radar points are more informative. This is particularly useful in environments with dense or occluded objects. BIRANet \cite{yadav_radar_2020} applies Concurrent Spatial and Channel Squeeze and Excitation (scSE) blocks for radar point clouds, using attention to emphasize radar-rich spatial areas. These scSE-enhanced features are fused with the camera data, significantly improving small-object detection. 

MVFusion \cite{wu_mvfusion_2023} uses radar and image data to produce dense BEV maps, relying on an attention mechanism to generate radar-guided and image-guided embeddings. Radar Transformer \cite{bai_radar_2021} utilizes vector and scalar attention mechanisms to implement spatial and Doppler radar information, preserving the range and elevation depth features unique to 4D radar. CramNet \cite{avidan_cramnet_2022} adapts to the additional depth provided by 4D radar with a ray-constrained cross-attention mechanism, projecting 3D radar information along image rays and refining depth estimates through radar data alignment.

Advancements in 4D imaging radar technology have enabled the generation of 3D point clouds. Existing methods often adapt elements from lidar-based models. Still, the point clouds generated by 4D imaging radars tend to be sparser and noisier than those from lidar, leading to reduced model performance. For example, some approaches apply PointPillars \cite{lang_pointpillars_2019} to 4D radar data, which works well due to pillarization being a good deterministic technique for handling sparse points. Other methods, like RPFA-Net \cite{xu_rpfa-net_2021}, improve orientation estimation by modifying PointPillars' pillarization process with a self-attention mechanism to capture global features. Similarly, SMURF \cite{liu_smurf_2024} adopts a multi-representation fusion strategy to simultaneously extract radar features using the PointPillars backbone and kernel density estimation (KDE) to generate enhanced feature maps. In LXL \cite{xiong_lxl_2023}, radar and camera features are projected into aligned voxel spaces using the SECOND \cite{yan_second_2018} backbone and are then fused within a dual-attention framework that emphasizes spatially relevant areas and feature details from each modality, refining features to yield robust object proposals.\\

%% file: method.tex
\section{Method}
\subsection{Overall Architecture}

\begin{figure*}[ht]
    \centering
    \includegraphics[width=\textwidth]{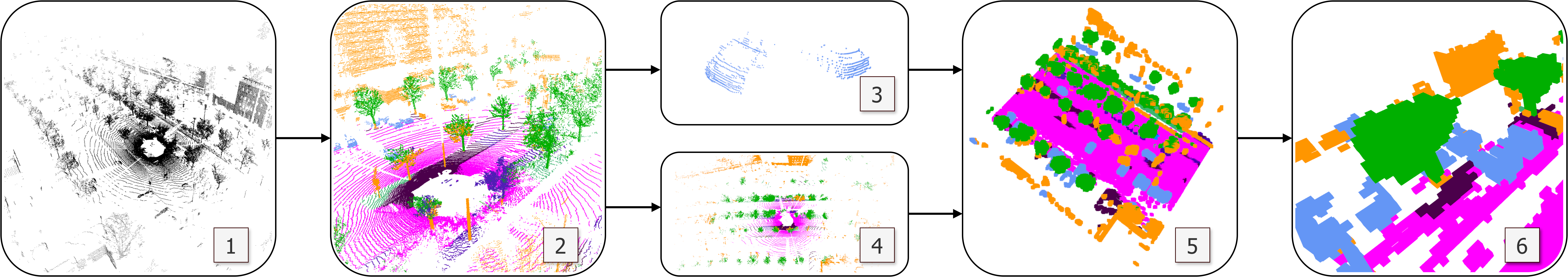}
    \caption{The process of auto-labeling dense occupancy pseudo-ground truth involves capturing dense lidar data (1), performing semantic segmentation (2), and separately extracting and accumulating dynamic (3) and static objects (4). Both point clouds are transformed into the world coordinate system (5) before being voxelized and refined to create the final occupancy labels (6).}
    \label{fig:auto-labeling}
\end{figure*}

The architecture of 4DRC-OCC, illustrated in Fig. \ref{fig:architecture}, is a robust dual-sensor framework designed to combine data from the camera and 4D radar, facilitating precise 3D semantic occupancy prediction. It employs two parallel branches, each optimized for its specific input source. The camera branch extracts features from monocular images and estimates depth, projecting these features into a 3D voxel grid. Simultaneously, the radar branch processes 4D radar point clouds to generate spatial data that complements the camera input. 
By leveraging the complementary strengths of these modalities, the synergy between the camera and radar branches in 4DRC-OCC enables the creation of a depth-aware 3D representation, which is crucial for robust and accurate occupancy predictions. The architecture is further enhanced with three distinct implementation variants:\\

\begin{itemize}
\item Version \textbf{A}: The vanilla implementation, which directly fuses 4D radar features and camera features within the 3D voxel space.\\
\item Version \textcolor{red_}{\textbf{B}}: Aimed at assisting the camera lifting mechanism, this variant builds upon \textbf{A} and enhances depth estimation by incorporating radar-derived depth cues into the camera input. Specifically, radar data is projected onto the image plane to create sparse pseudo-depth images.\\
\item Version \textcolor{blue_}{\textbf{C}}: Similar to Version B, this variant also enhances the camera lifting mechanism using radar-derived depth cues. However, it takes a different approach by directly integrating radar-derived depth information into the raw image to create RGB-D inputs, embedding depth information at the source.
\end{itemize}

Each Version reflects a unique strategy for combining radar and camera data, showcasing the flexibility of the 4DRC-OCC architecture. Version \textcolor{red_}{\textbf{B}} and Version \textcolor{blue_}{\textbf{C}} are part of the depth association (DA) strategies, aiming to improve the accuracy of camera depth estimation, which will be discussed.\\

\subsection{Camera Branch}
The camera branch employs multiple specialized components to extract, refine, and lift 2D features into a 3D voxel grid. Central to its functionality is the FB-BEV \cite{li_fb-bev_2023} architecture. To enhance the representation of depth information critical for accurate 3D projections, the camera branch includes a dedicated depth network supervised by lidar measurements. Additionally, a context network leverages camera intrinsics to capture spatial relationships and scene layout, ensuring alignment between the camera's perspective and the 3D voxel space.

The processing of camera images begins with a backbone network, which extracts rich feature representations from raw input images. These features are passed to a customized Feature Pyramid Network (FPN), consolidating multi-scale information and refining the representation. Unlike conventional FPNs, ours is tailored to produce a single-layer feature map with multiple channels optimized for semantic occupancy prediction. This feature map encapsulates critical scene details, serving as the foundation for further transformations into 3D space.

The transition from 2D image features to a 3D voxel grid is facilitated by a combination of depth estimation, spatial context encoding, and the Splatting module. A dedicated depth network predicts per-pixel depth distributions, which is crucial for accurately projecting 2D features into 3D space. This network divides the voxel space into 80 discrete, linear-spaced depth categories, enabling precise depth supervision during training. Ground-truth depth information from lidar is projected onto the image plane to supervise these predictions, ensuring alignment with the 3D voxel grid and enhancing depth estimation fidelity. Camera intrinsics are processed through the context network to generate spatially-aware scene representations. These features are combined with the 2D image features to create enriched, context-aware feature maps. The lifting mechanism splats \cite{hu_ea-lss_2023} the 2D image features onto the 3D voxel grid, utilizing the depth predictions and context features to map each pixel to its corresponding voxel location. This module builds on the principles of BEVDet \cite{huang_bevdet_2022} and BEVDepth \cite{li_bevdepth_2023}, but instead of compressing features into a BEV plane, these are directly processed in 3D voxel space. Features mapped to the same voxel are aggregated through sum pooling.\\

\subsection{Radar Branch}
The radar processing branch employs a two-stage approach leveraging the PointPillars \cite{lang_pointpillars_2019} and SECOND \cite{yan_second_2018} networks to convert radar point cloud data into a robust 3D feature representation, addressing the inherent sparsity and noise of 4D radar data. 
In the first stage, the PointPillars framework organizes the radar data into a grid of pillars in the x-y plane, with seven channels:

\begin{center}
    $ 
    \begin{Bmatrix} 
        \text{x} , 
        \text{y} , 
        \text{z} , 
        \text{velocity} , 
        \text{RCS} , 
        \text{confidence}
    \end{Bmatrix}$.  
\end{center}

The radar points within each non-empty pillar are processed collectively to generate a nine-dimensional feature vector for each point. These feature vectors are enhanced with statistical offsets to strengthen spatial and contextual information and passed through a simplified PointNet \cite{qi_pointnet_2017} module within PointPillars to yield a compact pseudo-image representation. After the PointPillars transformation, the pseudo-image output is processed by the SECOND backbone, a hierarchical convolutional network optimized for multi-scale spatial feature extraction. The SECOND backbone comprises three layers with increasing channels, enabling progressive downsampling and multi-scale feature capture. These multi-scale features are upsampled and concatenated across levels, creating a comprehensive representation that captures fine-grained details and larger spatial structures in BEV. The features are broadcasted to 3D voxel space by repeating the BEV maps along the height dimension.

\subsection{Fusion Mechanism}
The obtained multi-scale features from both branches are concatenated in voxel space and, after that, encoded by a 3D ResNet \cite{he_deep_2016} neck, which compresses and refines them into a unified single-scale voxel representation. This encoding ensures that the fused features are compact and expressive, enabling effective downstream processing. Finally, the unified voxel representation is passed to the occupancy head, a fully connected MLP. The occupancy head maps the features into an 18-channel probability distribution for each voxel, supporting detailed and probabilistic semantic predictions across the entire grid.

\subsection{Depth Association (DA)}
Camera image depth association addresses the challenges of monocular depth estimation by leveraging the elevation and spatial precision of 4D radar. Two refined strategies enhance the base architecture introduced in Version A.
In Version B, radar point clouds are projected onto the image plane without embedding radar depth directly within the RGB channels. Instead, sparse pseudo-images are generated from the radar data and aligned with downsampled RGB images. These pseudo-images are concatenated with camera features at the feature level, preserving radar depth as a separate modality. A custom CNN processes the concatenated features, enriching the overall representation by treating radar and RGB data as complementary sources while enhancing depth awareness.
In Version C, radar point clouds are projected directly onto the image plane, with radar-derived depth values added as an additional channel at corresponding pixel locations to create sparse RGB-D images. To comply with common architectures and image backbones, an additional CNN is employed to transform these RGB-D images back into a standard RGB format. This transformation ensures compatibility with pre-trained RGB networks, resulting in a distorted image that is not visually meaningful to humans. This approach maintains architectural consistency with Version A while seamlessly integrating radar depth cues into the network.

\subsection{Auto-labeling Pseudo Grund Truth}
The process of generating dense occupancy pseudo ground truth is illustrated in Fig. \ref{fig:auto-labeling}. Our approach draws inspiration from SurroundOcc \cite{wei_surroundocc_2023}, which utilizes multi-frame lidar data combined with 3D detection and semantic segmentation labels to eliminate the need for manual annotation. In SurroundOcc, multi-frame lidar point clouds are first stitched separately for static scenes and dynamic objects by extracting and transforming points from each frame into a unified world coordinate system. These points are then concatenated to form the complete scene. The point cloud is densified using Poisson Surface Reconstruction \cite{kazhdan_poisson_nodate}, filling gaps by creating a triangular mesh and evenly distributing vertices, thus improving point density. Semantic labels are then assigned to the dense voxel representation via a Nearest Neighbors (NN) algorithm, matching the occupied voxels with the nearest semantic labels from sparse voxelized data. However, unlike SurroundOcc, our approach bypasses the computationally expensive and potentially inaccurate Poisson Surface Reconstruction. With the dense point clouds provided by a 128-beam lidar sensor, we eliminate the need for densification techniques. The inherent density of the lidar data ensures sufficient coverage, allowing us to directly generate high-quality volumetric occupancy data without introducing artificial points or requiring interpolation. Point-wise semantic labels are obtained using a pre-trained PointTransformerV3 \cite{wu_point_2024} model trained on the nuScenes \cite{caesar_nuscenes_2020} dataset, enabling precise occupancy labeling with minimal post-processing. This approach demonstrates that dense lidar data and advanced point-wise semantic labeling models can achieve high-quality occupancy ground truth without requiring additional computational overhead. To further enhance label consistency, we mitigate noise from the auto-labeling process by reassigning the class of lonely voxels, spatially isolated from neighboring voxels of the same class, by matching their class to that of neighboring voxels. This step improves the accuracy and consistency of the semantic occupancy labels, ensuring high-quality ground truth with minimal manual intervention. Voxel labeling is determined based on the interaction between the voxel and lidar points. If a voxel reflects a lidar point, it is assigned the same semantic label as the lidar point. If a lidar beam passes through a voxel in the air, the voxel is classified as free. 
Our dataset, Perciv-scenes, contains approximately 30K samples and is structured in line with the settings used by the Occ3D-nuScenes dataset \cite{tian_occ3d_2023}: occupancy annotations are provided for each frame within the spatial bounds of [-40, -40, 1, 40, 40, 5.4] meters, with a voxel resolution of 0.4 meters, and includes 18 semantic classes.
The camera used in our setup is an IDS uEye, capturing colored, rectified images with a resolution of 1936 × 1216 pixels at approximately 30 Hz. The camera's horizontal field of view is 64° (±32°), and its vertical field of view is 44° (±22°). The lidar sensor is an Ouster 128, mounted on the vehicle's top and operating at 20 Hz. The lidar point clouds are ego-motion compensated to account for both the vehicle's motion during the scan and the motion between the capture of lidar and camera data. Additionally, the radar sensor is a ZF FRGen21 4D radar, operating at 13 Hz and mounted behind the front bumper. The radar point clouds are also ego-motion compensated to account for motion between radar and camera data capture.

\subsection{Loss Functions}
The training relies on a multi-faceted loss function that ensures accuracy in-depth, occupancy, and semantic prediction. We employ a distance-aware Focal loss \( L_{fl} \) inspired by M2BEV \cite{xie_m2bev_nodate} to prioritize hard-to-classify instances. Additionally, Dice loss \( L_{dl} \), Semantic Affinity losses \( L_{geo scal} \) and \( L_{sem scal} \), derived from MonoScene \cite{cao_monoscene_2022}, are used to capture detailed semantic structures. Finally, the Lovasz-Softmax loss \( L_{ls} \), based on OpenOccupancy \cite{wang_openoccupancy_2023} provides fine-grained boundary accuracy. Depth supervision loss \( L_d \) and a 2D semantic loss \( L_s \) are also incorporated to supervise depth prediction and semantic consistency.

%% file: experiments.tex
\section{experiments}

\input{tables/eval_18_classes}

\subsection{Implementation Details}
We train the Baseline and Versions A, B, and C both from scratch, as well as started from a FB-OCC \cite{li_fb-occ_2023} checkpoint to fine-tune the camera branch only, in Table \ref{tab:overall} marked with a trailing '-ft.' The training setup utilizes a single Nvidia RTX4090 GPU with a batch size of 6 samples spanning 20  epochs. The model leverages a pre-trained ResNet50 \cite{he_deep_2016} for image backbone processing, which is updated during training, except for the first layer, which is frozen. The input images are scaled to 256$\times$704 and image features down sampled with a stride of 16, spanning 256 channels. The depth prediction network follows a similar configuration for the image input. After image feature lifting, the multi-scale features will have [256, 128, 64] channels. This is also the case for 4D radars, with the channel count doubling after the concatenation of both modalities.
Our dataset only includes front-facing camera images and radar data, so we train exclusively on occupancy labels within this field of view. To support occlusion reasoning, we retain all ground-truth voxels during training, including those not visible in the camera view. Leveraging radar data in addition to camera input enhances the ability to reason about occlusions, as 4D radar can detect objects and spatial occupancy in regions obscured from the line of sight of the camera. This is especially valuable in complex environments or scenarios with limited visibility, where radars can penetrate obstructions, helping the model achieve a more robust spatial understanding.
 
\subsection{Overall performance}

\begin{figure*}[ht]
    \centering
    \includegraphics[width=\textwidth]{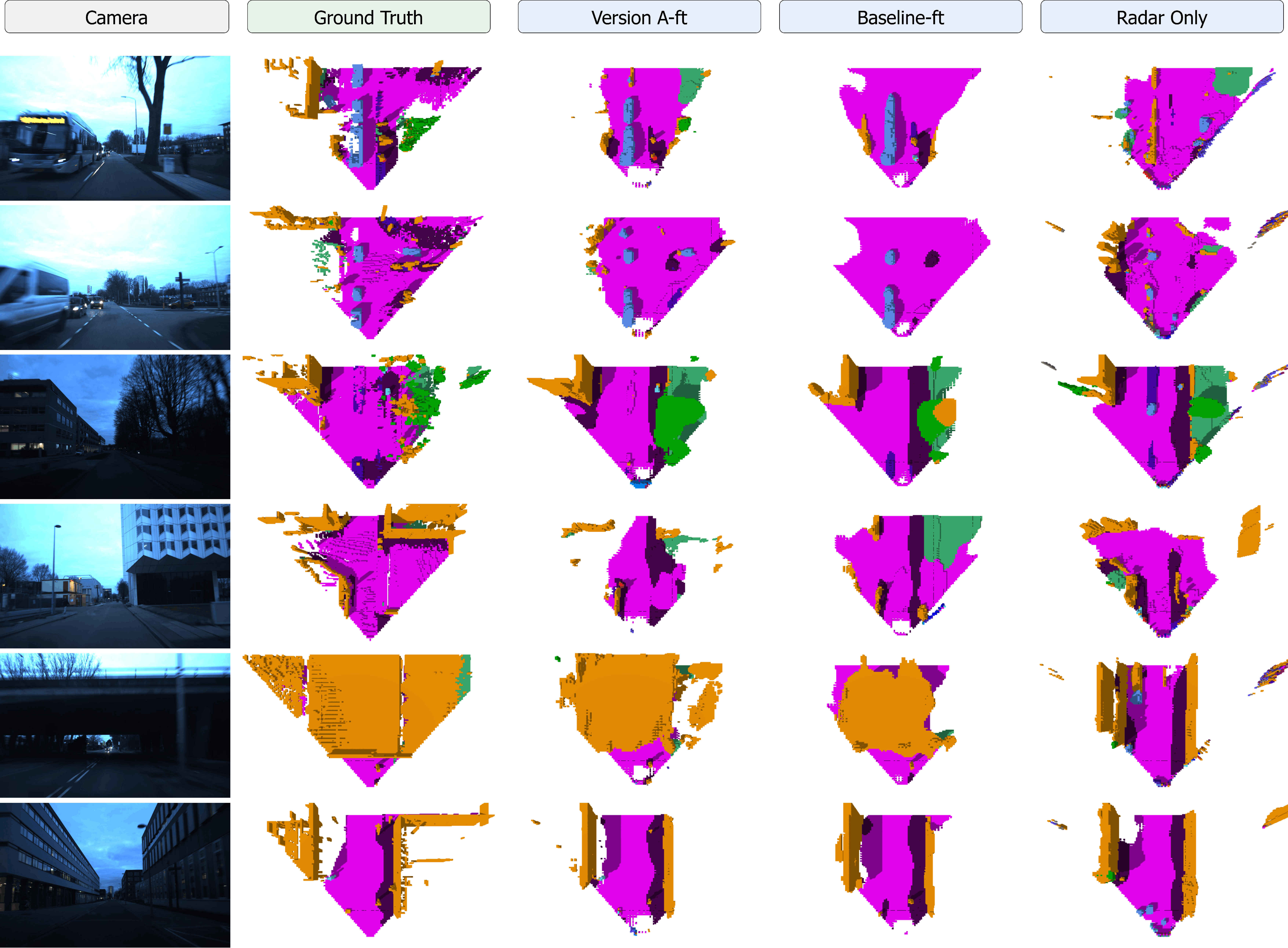}
    \caption{4DRC-OCC improves prediction accuracy in poor lighting conditions by integrating data from 4D radar and cameras.}
    \label{fig:example_preds-modality}
\end{figure*}

To evaluate the contribution of 4D radar, we use the camera branch as the Baseline for performance comparison, providing a clear metric to assess the improvements introduced by integrating 4D radar data. The results are summarized in Table \ref{tab:overall}, with additional network variants included to provide a comprehensive overview. These are explained in detail in the ablation study. A visual comparison is presented in Fig. \ref{fig:example_preds-modality}, with the interpretation of this figure left to the reader. The evaluation of our model is based on the Mean Intersection over Union (mIoU), defined as:

\begin{equation}
    mIoU = \frac{1}{C} \sum^{C}_{c=1} \frac{\textit{TP}_c}{\textit{TP}_c + \textit{FP}_c + \textit{FN}_c},
\end{equation}
where $\text{TP}_c$, $\text{FP}_c$, and $\text{FN}_c$ denote the number of true positive, false positive, and false negative predictions for class $c$, respectively, and $C$ represents the total number of classes.\\

Given the substantial class imbalance in our dataset, we also report the weighted mIoU alongside the standard mIoU. The weighted mIoU considers the relative frequency of each class in the dataset, assigning higher weights to classes that appear more frequently. This approach provides a more balanced evaluation of model performance by emphasizing the scores of classes with more training data, ensuring that the performance of these classes has a proportionate impact on the overall score. The frequencies of the classes, represented as percentages, are shown in Table \ref{tab:overall}. Note that some classes might not be in our dataset but are listed to adhere to convention and ensure consistency with other research.

Our approach achieves significant improvements in both mIoU and weighted mIoU metrics. The radar-enhanced models show increased accuracy across nearly all semantic categories compared to the Baseline models. Notably, Version B and Version C achieved the highest mIoU scores, with \scrB\ and \ftC\ reaching a top mIoU of 17.3, a 36\% improvement over \ftbaseline. Furthermore, the weighted mIoU peaked at 32.7 in \scrB. Notably, Versions B and C outperformed Version A, demonstrating that the depth association strategies employed in these models deliver substantial benefits.

The best-performing scores for each semantic class are consistently observed within the radar-integrated models. Classes such as bicycle, pedestrian, and car, often challenging to detect due to their small data representation, saw significant improvements. These results underscore the potential of radar-camera fusion to enhance detection and classification accuracy across diverse semantic categories, particularly for classes that are difficult to predict using monocular vision alone.

Version A and Version C benefit from fine-tuning, while Version B does not significantly improve. In Version A, fine-tuning refines the fusion of radar and camera features in the 3D voxel space, enhancing the alignment and integration of depth cues and semantic information, leading to more accurate occupancy predictions. In Version C, where radar-derived depth is directly embedded into the RGB images, fine-tuning helps the model better process these RGB-D inputs, improving depth awareness and prediction accuracy. In contrast, Version B, which uses sparse pseudo-depth images created by projecting radar data onto the image plane, does not benefit as much from fine-tuning. The radar depth is treated separately from the RGB features, and its sparse representation does not provide enough structured depth cues for fine-tuning to improve depth integration and occupancy prediction.  These findings underscore the powerful integration of 4D radar into monocular semantic occupancy prediction, enabling more robust and precise scene understanding.

\begin{figure*}[ht]
    \centering
    \includegraphics[width=\textwidth]{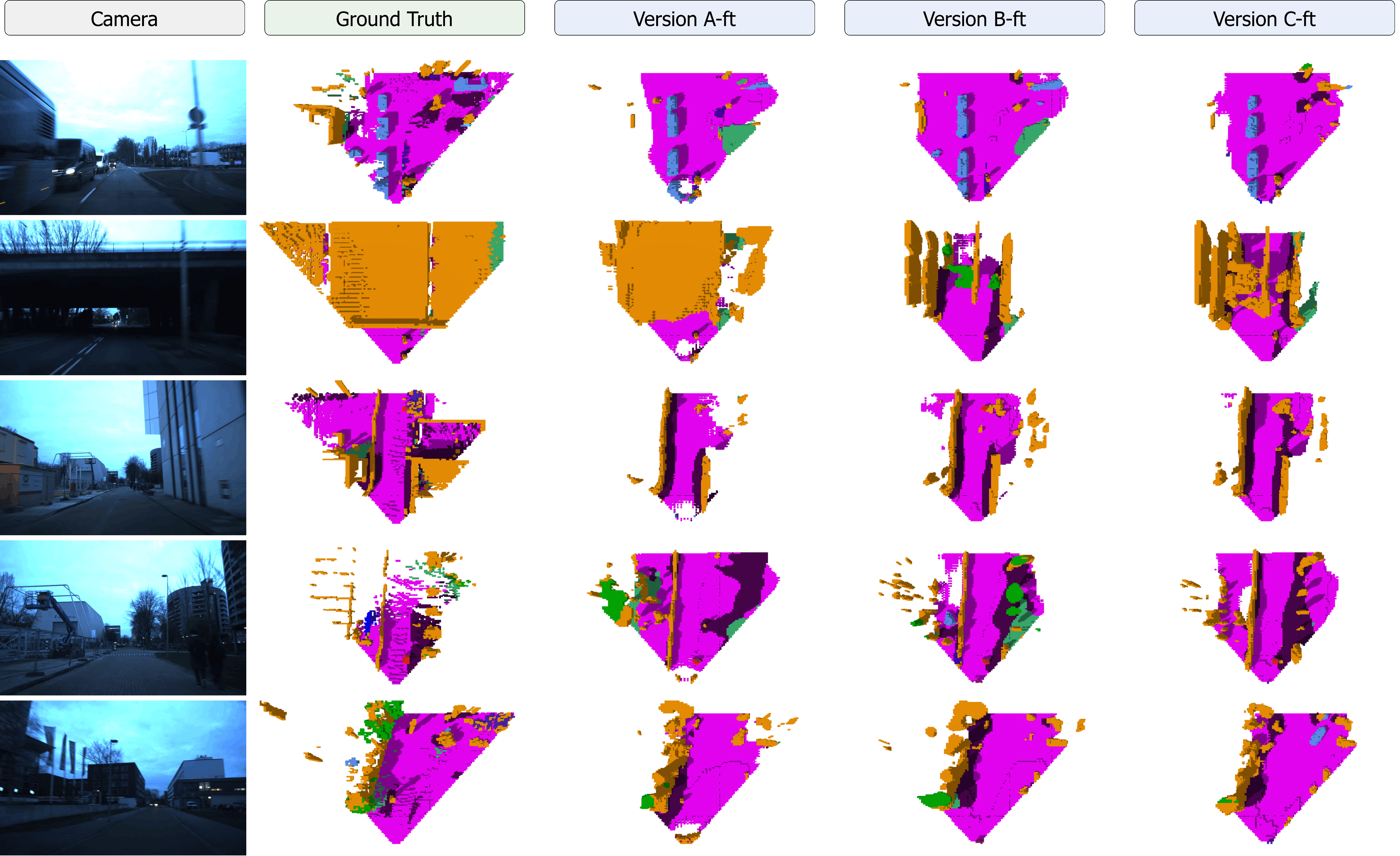}
    \caption{Depth association strategies enhance predictions by improving spatial accuracy and semantic understanding in complex environments.}
    \label{fig:example_preds-depthass}
\end{figure*}

\subsection{Occupancy}
We access the geometric accuracy of the networks by merging all semantic classes into a single class presented in Table \ref{tab:occupancy}. This approach simplifies identifying any occupied space, providing insight into the overall spatial awareness.

\input{tables/eval_1_class}

The results demonstrate that versions B and C deliver the highest accuracy, with \scrC\ achieving the best mIoU at 44.7\%, followed closely by \ftC\ and \scrB. This suggests that the depth association strategies link radar depth cues directly with image features and are highly effective for occupancy prediction, underscoring the benefits of 4D radar-camera fusion for reliable geometric awareness.

\subsection{Ablation Study}
 We conduct two ablation studies to analyze the contributions of radar data to monocular semantic occupancy prediction. The first study involves training the network on radar data only, providing insight into the complementary value of radar for these networks. The second study examines the added value of 4D radar by training the network on radar data without elevation information, mimicking conventional 3D radar systems.

In the first ablation, the Radar Only network in Table \ref{tab:overall} achieves notable performance, scoring 10.4 for the bicycle class. While \ftbaseline achieves 12.2 for the same class, the fusion \scrC outperforms both with a score of 29.4, demonstrating the synergistic advantage of combining radar and camera data. This trend is consistent across other semantic classes, where the performance of the fusion networks exceeds the sum of the individual radar and camera branches. Dynamic classes, such as car and pedestrian, show remarkable improvements with fusion networks, highlighting the complementary nature of radar and camera modalities in enhancing overall detection accuracy. Visualizations of these predictions are shown in Fig. \ref{fig:example_preds-modality}.

In the second ablation, we compare the performance of models trained with full 4D radar (\scrA\ and \ftA) to those trained without elevation information (\scrAnoelev\ and \ftAnoelev), representing conventional 3D radar. The results demonstrate the added value of 4D radar elevation information in the features: \ftA\ achieves a mIoU of 15.8, outperforming \ftAnoelev\ with a mIoU of 14.2, reflecting an 11.3\% improvement. Similarly, \scrA\ achieves a mIoU of 15.5, surpassing \scrAnoelev\ with a mIoU of 14.5, corresponding to a 6.9\% improvement. These results highlight the contribution of elevation information in improving network performance by providing richer spatial features.

\subsection{Limitations and Future Work}
In this work, several limitations are associated with the chosen approach. We highlight four key areas where improvements can be made: the fusion mechanism, the quality of the ground truth, and the class imbalance in the dataset. These limitations are discussed in greater detail below, along with potential solutions.

Our method of fusing radar and camera features relies on a simple concatenation approach. While this basic method successfully merges the sensor modalities, it does not fully capture the complementary and nuanced nature of the inputs, which differ significantly in terms of spatial and temporal resolution and the type of information they provide. A more advanced fusion strategy, such as one employing attention mechanisms, could enable the model to better assess the relevance of each modality based on the context. This would be especially important for accurate semantic occupancy prediction. For example, radar data might be more reliable in scenarios with poor weather or low visibility than camera data, and a sophisticated fusion mechanism would allow the model to prioritize radar features in such conditions \cite{wang_unibev_2024}. This would be evident in the occupancy predictions, where areas impacted by reduced visibility could still yield accurate results due to the radar’s more dependable input. A more complex fusion method would thus improve the blending of these features, enhancing semantic segmentation quality and leading to more robust, context-aware predictions. 
For future work, incorporating an attention mechanism to fuse the features of both modalities could significantly improve the model’s performance. However, this would be challenging due to computational limitations, mainly when dealing with multi-scale voxel features that need to be merged. Fusing such high-dimensional data requires substantial computational resources, which could limit the practicality of this approach. Therefore, more research is needed to explore computationally efficient strategies for feature fusion and spatially effective data representations. Recent works \cite{huang_tri-perspective_2023, ma_cotr_2023, lu_octreeocc_2023, yan_sparse_2021}, have made significant progress in this area, offering potential solutions for merging multi-modal data in a more computationally feasible and efficient manner while preserving spatial information quality.

Another limitation is the absence of dropout training, which hampers the model's ability to handle missing or degraded sensor inputs. Implementing dropout during training could increase robustness by enabling the network to operate effectively even when one of the sensors experiences partial or complete failure. For example, if the radar or camera becomes temporarily unavailable, a dropout-trained model would be better equipped to maintain accurate occupancy predictions based on whichever sensor remains functional. Without dropout, the current model will likely struggle with adaptation, resulting in reduced performance or erratic predictions.

The quality of the ground truth used for training the semantic occupancy prediction model is also a significant challenge. Despite efforts to obtain accurate annotations, the ground truth remains noisy due to inconsistencies in labeling, sensor inaccuracies, and ambiguities in semantic classification. Additionally, there is a considerable class imbalance in the dataset, with certain classes being underrepresented compared to others. This imbalance exacerbates the issue, as the model focuses more on the dominant classes while neglecting the minority ones. The combination of noisy ground truth and class imbalance makes it difficult for the model to learn accurate representations for all classes, resulting in suboptimal performance. In particular, the model struggles to generalize well for the underrepresented classes, leading to lower overall prediction accuracy. These challenges hinder the model's performance and limit its applicability in real-world scenarios, where balanced and high-quality ground truth data are often difficult to obtain.

Improving the auto-labeling of occupancy ground truth for future work could be crucial in addressing these challenges. One promising approach would be implementing a tracker for bounding boxes that could assist in more accurately segregating dynamic objects. This method, as demonstrated in Fig. \ref{fig:auto-labeling}, could help ensure that moving objects are consistently tracked and labeled, reducing the impact of noisy annotations and improving the accuracy of segmentation in dynamic environments \cite{liu_offline_2024}. Another potential avenue for improvement is implementing a Bayesian update for voxel-based point cloud aggregation and voxelization. This approach would enable probability reasoning about the semantic content of each voxel, allowing the model to incorporate uncertainty in the data and improve the robustness of semantic predictions \cite{gebraad_leap_nodate}. These advancements would enhance the overall quality of the ground truth.

%% file: tables/eval_18_classes.tex
\renewcommand{\arraystretch}{1.2}

\begin{table*}[ht]
    \centering
    \footnotesize
    \begin{tabular}{l|c|c|ccccccccccccccccc}
    \hline
        \centering
        Version &
        \rotatebox{90}{\textbf{mIoU}} & 
        \rotatebox{90}{\textbf{weighted mIoU}} & 
        \rotatebox{90}{\textcolor{others}{\rule{2mm}{2mm}} others (0.00\%)} &
        \rotatebox{90}{\textcolor{barrier}{\rule{2mm}{2mm}} barrier (0.03\%)} &
        \rotatebox{90}{\textcolor{bicycle}{\rule{2mm}{2mm}} bicycle (0.15\%)} &
        \rotatebox{90}{\textcolor{bus}{\rule{2mm}{2mm}} bus (0.04\%)} &
        \rotatebox{90}{\textcolor{car}{\rule{2mm}{2mm}} car (2.82\%)} &
        \rotatebox{90}{\textcolor{construction_vehicle}{\rule{2mm}{2mm}} constr. veh. (0.02\%) } &
        \rotatebox{90}{\textcolor{motorcycle}{\rule{2mm}{2mm}} motorcycle (0.01\%)} &
        \rotatebox{90}{\textcolor{pedestrian}{\rule{2mm}{2mm}} pedestrian (0.10\%)} &
        \rotatebox{90}{\textcolor{traffic_cone}{\rule{2mm}{2mm}} traffic cone (0.00\%)} &
        \rotatebox{90}{\textcolor{trailer}{\rule{2mm}{2mm}} trailer (0.05\%)} &
        \rotatebox{90}{\textcolor{truck}{\rule{2mm}{2mm}} truck (0.51\%)} &
        \rotatebox{90}{\textcolor{driveable_surface}{\rule{2mm}{2mm}} driv. surf. (28.45\%)} &
        \rotatebox{90}{\textcolor{other_flat}{\rule{2mm}{2mm}} other flat (0.89\%)} &
        \rotatebox{90}{\textcolor{sidewalk}{\rule{2mm}{2mm}} sidewalk (5.70\%)} &
        \rotatebox{90}{\textcolor{terrain}{\rule{2mm}{2mm}} terrain (3.11\%)} &
        \rotatebox{90}{\textcolor{manmade}{\rule{2mm}{2mm}} manmade (40.81\%)} &
        \rotatebox{90}{\textcolor{vegetation}{\rule{2mm}{2mm}} vegetation (17.32\%)} \\
        \hline
        \scrRO          & 11.3          & 26.6          & - & 0.0          & 10.4          & 5.6           & 17.9          & \textbf{3.3} & 0.0          & 0.3           & - & 1.3          & 5.1          & 49.0          & 10.6          & 19.1          & 8.1           & 17.2          & 21.1          \\
        \scrbaseline    & 11.7          & 24.2          & - & \textbf{1.5} & 12.2          & 3.6           & 23.7          & 0.2          & 0.1          & 7.1           & - & 3.3          & 7.0          & 48.6          & 10.8          & 16.0          & 12.2          & 13.5          & 16.0          \\
        \ftbaseline     & 12.1          & 24.6          & - & 0.0          & 12.6          & 5.4           & 27.0          & 1.2          & 0.1          & 7.5           & - & 2.4          & 7.2          & 48.7          & 8.3           & 21.3          & 10.1          & 14.2          & 14.9          \\
        \ftAnoelev  & 14.2          & 31.0          & - & 0.0          & 15.7          & 4.0           & 31.1          & 2.2          & 0.0          & 11.0          & - & 1.0          & 5.7          & 52.4          & 10.7          & 21.0          & 13.3          & 23.7          & 22.0          \\
        \scrAnoelev & 14.5          & 30.0          & - & 0.7          & 15.2          & 3.2           & 27.9          & 0.7          & 1.5          & 14.9          & - & \textbf{3.8} & 9.4          & 51.0          & 9.6           & 20.7          & 13.3          & 22.1          & 22.7          \\
        \scrA           & 15.5          & 31.2          & - & 0.0          & 23.0          & 5.9           & 31.0          & 0.2          & 0.5          & 12.0          & - & 2.7          & 9.6          & 52.7          & 12.6          & 21.6          & 14.1          & 22.8          & 23.7          \\
        \scrC           & 15.6          & 32.4          & - & 0.0          & 25.7          & 7.2           & 32.8          & 3.2          & 0.0          & 12.1          & - & 0.6          & 6.2          & 53.6          & 9.5           & 22.1          & 12.6          & \textbf{25.3} & 23.2          \\
        \ftA            & 15.8          & 31.1          & - & 0.0          & 22.9          & 10.7          & 32.9          & 1.6          & 0.0          & 15.1          & - & 3.7          & \textbf{9.6} & 52.8          & 7.2           & 21.6          & 13.5          & 22.6          & 23.5          \\
        \ftB            & 16.7          & 31.8          & - & 0.0          & 27.5          & \textbf{13.2} & 33.1          & 1.2          & 0.4          & 14.6          & - & 2.0          & 8.0          & 53.8          & 14.5          & 23.4          & 12.5          & 23.2          & 23.9          \\
        \ftC            & \textbf{17.3} & 32.3          & - & 0.0          & \textbf{29.4} & 11.4          & \textbf{33.9} & 2.0          & \textbf{2.6} & \textbf{17.8} & - & 1.7          & 9.1          & 53.1          & 12.7          & \textbf{24.3} & 14.0          & 25.0          & 23.2          \\
        \textbf{\scrB}  & \textbf{17.3} & \textbf{32.7} & - & 0.3          & 29.1          & 13.1          & 33.5          & 1.9          & 0.0          & 14.4          & - & 2.7          & 8.5          & \textbf{54.2} & \textbf{14.7} & 23.8          & \textbf{14.5} & 24.2          & \textbf{25.3} \\
        \hline
    \end{tabular}
    \vspace{2mm}
    \caption{Semantic occupancy prediction performance on the Perciv-scenes dataset.}
    \label{tab:overall}
\end{table*}

%% file: tables/eval_1_class.tex
\begin{table}[ht]
    \centering
    \footnotesize
    \begin{tabular}{l|c}
    \hline
        \centering
        Version & 
        \rotatebox{90}{\textcolor{others}{\rule{2mm}{2mm}} occupied } \\
        \hline
        \scrbaseline    & 35.1          \\
        \ftbaseline     & 35.5          \\
        \scrRO          & 39.1          \\
        \scrAnoelev & 41.0          \\
        \scrA           & 42.3          \\
        \ftAnoelev  & 42.7          \\
        \ftA            & 42.8          \\
        \ftB            & 43.8          \\
        \ftC            & 44.2          \\
        \scrB           & 44.5          \\
        \textbf{\scrC}  & \textbf{44.7} \\
        \hline
    \end{tabular}
    \vspace{2mm}
    \caption{Occupancy performance on the perciv-scenes dataset.}
    \label{tab:occupancy}
\end{table}

%% file: conclusion.tex
\section{Conclusion}

This study presents a pioneering 3D semantic occupancy prediction approach by integrating 4D radar data with monocular camera input. The results demonstrate that 4D radar serves as a robust complementary sensor, significantly improving the accuracy of occupancy predictions, particularly under challenging conditions such as adverse weather and poor lighting. Including radar depth and velocity information, alongside the high-resolution camera features, leads to more reliable and accurate semantic segmentation, overcoming many limitations inherent in monocular vision systems. Using an automatically labeled dataset further enhances the scalability of the proposed framework, reducing the need for manual annotation and enabling more efficient training.
The proposed 4DRC-OCC architecture, through its multi-scale feature fusion and depth association strategies, highlights the potential of radar-camera fusion in enhancing depth estimation and semantic occupancy accuracy. Notably, the radar-integrated models outperformed monocular baselines, especially in detecting challenging object classes such as bicycles, pedestrians, and static obstacles. The findings underscore the importance of sensor fusion in addressing the limitations of individual sensors, paving the way for more resilient and contextually aware autonomous vehicle perception systems.his work identifies key limitations, including a simple fusion mechanism, noisy ground truth, and lack of dropout training. Future improvements could involve using advanced fusion techniques like attention mechanisms, adding dropout for robustness, and enhancing ground truth quality. These enhancements would improve model performance and resilience in real-world applications.

%% file: acknowledgement.tex
\section*{Acknowledgment}

I am deeply grateful to those who supported and guided me throughout this research. I sincerely thank Perciv AI, particularly Andras, for their invaluable insights and for fostering a collaborative environment that enriched my analysis. I also sincerely thank my academic advisor, Holger, for his expert guidance and constructive feedback, which were vital to the success of this paper. Finally, I acknowledge the foundational work of the authors whose papers informed this research.

\begin{itemize}
    \item PointTransformerV3 \cite{wu_point_2024}
    \item SurroundOcc \cite{wei_surroundocc_2023}
    \item PointPillars \cite{lang_pointpillars_2019}
    \item FB-BEV \cite{li_fb-occ_2023}\\
\end{itemize}

Thank you all for your valuable contributions to my academic journey.